\pdfoutput=1

\documentclass[11pt]{article}

\usepackage[final]{acl}

\usepackage{times}
\usepackage{latexsym}

\usepackage[T1]{fontenc}

\usepackage[utf8]{inputenc}

\usepackage{microtype}

\usepackage{inconsolata}

\usepackage{graphicx}
\usepackage{amsmath,amssymb}
\usepackage{booktabs}       
\usepackage{multirow}       
\usepackage{array}          
\usepackage{pgfplots}
\usepackage{tikz}        
\usepackage{subcaption}  
\pgfplotsset{compat=1.18}
\usetikzlibrary{patterns}
\usepackage{algorithm}
\usepackage{algorithmic}
\usepackage{tcolorbox}
\tcbuselibrary{breakable}

\title{LegalReasoner: Step-wised Verification-Correction \\ for Legal Judgment Reasoning}

\author{
 \textbf{Weijie Shi\textsuperscript{1}\thanks{Co-authors: Han Zhu}\thanks{\small{
   \textbf{Email:} \href{mailto:wshiah@connect.ust.hk}{wshiah@connect.ust.hk}
 }}},
 \textbf{Han Zhu\textsuperscript{1}\footnotemark[1]},
 \textbf{Jiaming Ji\textsuperscript{2}},
 \textbf{Mengze Li\textsuperscript{1}},
\\
 \textbf{Jipeng Zhang\textsuperscript{1}},
 \textbf{Ruiyuan Zhang\textsuperscript{1}},
 \textbf{Jia Zhu\textsuperscript{3}},
 \textbf{Jiajie Xu\textsuperscript{4}},
\\
 \textbf{Sirui Han\textsuperscript{1}\thanks{Corresponding authors: Sirui Han, Yike Guo}},
 \textbf{Yike Guo\textsuperscript{1}\footnotemark[3]}
\\
\\
 \textsuperscript{1}The Hong Kong University of Science and Technology,
\\
 \textsuperscript{2}Peking University,
\\
 \textsuperscript{3}Zhejiang Key Laboratory of Intelligent Education Technology and Application, Zhejiang Normal University,
\\
 \textsuperscript{4}Soochow University
}

\begin{document}
\maketitle
\begin{abstract}
Legal judgment prediction (LJP) aims to function as a judge by making final rulings based on case claims and facts, which plays a vital role in the judicial domain for supporting court decision-making and improving judicial efficiency. However, existing methods often struggle with logical errors when conducting complex legal reasoning. We propose LegalReasoner, which enhances LJP reliability through step-wise verification and correction of the reasoning process. Specifically, it first identifies dispute points to decompose complex cases, and then conducts step-wise reasoning while employing a process verifier to validate each step's logic from correctness, progressiveness, and potential perspectives. When errors are detected, expert-designed attribution and resolution strategies are applied for correction. To fine-tune LegalReasoner, we release the LegalHK dataset, containing 58,130 Hong Kong court cases with detailed annotations of dispute points, step-by-step reasoning chains, and process verification labels. Experiments demonstrate that LegalReasoner significantly improves concordance with court decisions from 72.37 to 80.27 on LLAMA-3.1-70B. The data is available at https://huggingface.co/datasets/weijiezz/LegalHK.
\end{abstract}

\section{Introduction}
Legal judgment prediction (LJP) aims to function as a judge by predicting rulings based on facts of legal cases. It serves crucial roles in supporting legal decision-making processes, improving judicial efficiency and justice \cite{cui2023survey}, and enabling citizens to grasp potential case outcomes without expensive legal consultation \cite{feng2022legal}. However, when confronted with complex cases, LJP requires a highly specialized ability to capture key points of cases and conduct a thorough analysis to reach reliable and well-founded conclusions.

Unfortunately, existing LJP methods overlook the importance of logical reasoning in judicial decision-making. While BERT-based models like LegalBERT \cite{chalkidis2019neural, chalkidis2020legal} and jurBERT \cite{masala2024improving} have demonstrated promising results, they inherently lack logical reasoning capabilities. Although large language models (LLMs) have shown significant improvements in executing complex multi-step reasoning through carefully designed prompts \cite{trautmann2022legal}, retrieval-augmented generation \cite{santosh2024incorporating}, and fine-tuning \cite{ammar2024prediction}, even state-of-the-art LLMs still frequently exhibit logical errors. To remedy the reasoning vulnerabilities of LLMs in the LJP task, particularly when dealing with high-complexity cases, there is a critical need for both guided reasoning and robust verification mechanisms that can detect and rectify logical errors throughout the inference process.

To enable effective legal reasoning, there are a series of challenges to maintain logical consistency and generate well-founded judgments. First, legal cases often involve implicit causal relations and contentions, making them more complex. Second, the lack of effective process supervisors in judicial reasoning makes it challenging to monitor and validate the logical flow of legal analysis, potentially leading to errors and inconsistent conclusions. Third, addressing reasoning errors requires problem attribution and solutions, with each type of error demanding a tailored remediation approach. For instance, when legal provisions are misunderstood, we need to review the relevant laws for clarity. Similarly, when reasoning doesn't match case facts, we must check that the logic aligns with the evidence.

\begin{figure*}[t]
\centering
\begin{minipage}[t]{1\linewidth}
\centering
\includegraphics[width=1.0\textwidth]{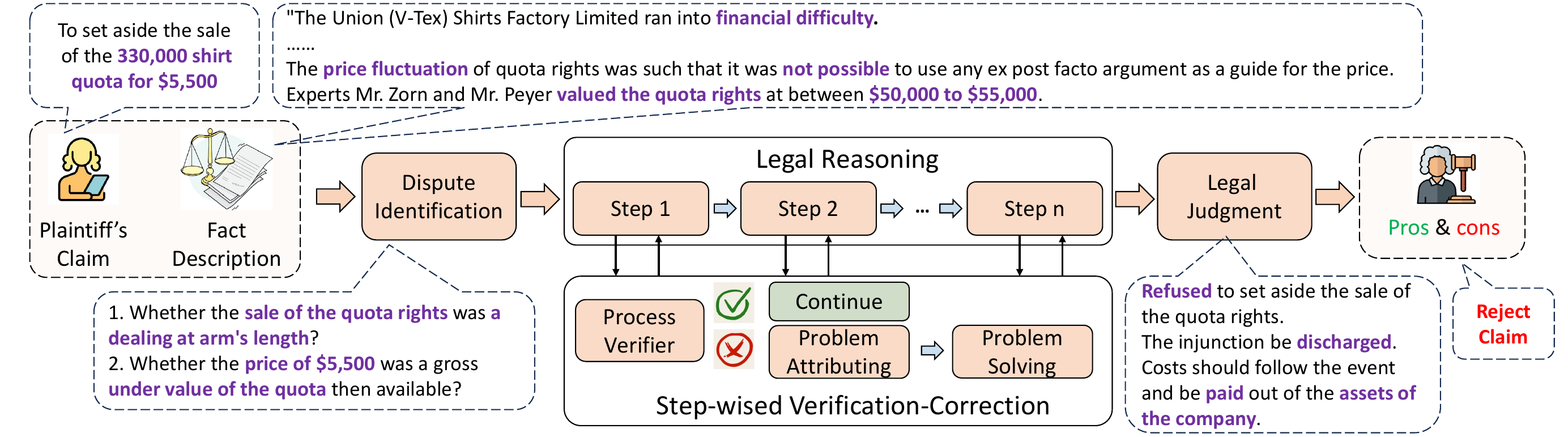}
\end{minipage}
\centering
\caption{An illustration of legal judgment prediction. LegalReasoner first breaks down the process into identifying disputes, and then subsequent reasoning steps are monitored by a verifier. If errors occur, they are traced back to predefined issues and corrected by tailored strategies, ensuring reliable judicial reasoning.}
\label{fig:architecture}
\end{figure*}

In this paper, we propose LegalReasoner, which first identifies the disputes to decompose complex cases, and then step-wise verifies and corrects potential errors in the legal reasoning process as depicted in Figure \ref{fig:architecture}. To facilitate legal case decomposition and reasoning, we construct LegalHK, a comprehensive dataset containing 58,130 Hong Kong court cases, which are annotated with dispute points, evidence analysis, relevant statutory provisions, precedent citations, and step-by-step reasoning chains leading to final judgments. Upon this foundation, a process verifier is proposed to validate each legal reasoning step. When reasoning errors are detected, our correct component first attributes to specific error types designed by legal experts, such as legal principle misapplication, discrepancy between facts and reasoning, etc. Next, it activates targeted problem-solving mechanisms to address these issues through tailored strategies, like provision retrieval, fact tracking, enforcement pathway analysis, etc. Our key contributions can be summarized as follows:
\begin{itemize}
    \item To ensure the reliability of judicial reasoning, we propose a process verifier to detect errors at each step, complemented by expert-designed attribution and resolution strategies to correct errors.
    \item To enhance case decomposition and structured reasoning in LLMs, we open-source the LegalHK dataset with detailed annotations.
    \item To the best of our knowledge, we are the first to propose verification and correction mechanisms for the legal reasoning process, improving concordance with court decisions from 72.37 to 80.27 on LLAMA-3.1-70B.
\end{itemize}

\section{Related Work}
\subsection{Legal Judgment Prediction}
Existing Legal judgment prediction (LJP) methods primarily fall into two categories: BERT-based and LLM-based approaches.
BERT-based methods treat the LJP task as a multi-class classification to identify appropriate charges \cite{huang2024cmdl,de2022explainable}. To enhance legal knowledge, many researchers \cite{chalkidis2019neural,chalkidis2020legal} focused on fine-tuning by legal corpora. \citet{feng-etal-2022-legal} and \citet{deroy2024ensemble} supplied judgment information by constrained event extraction and document summarization, respectively. \citet{masala2024improving} introduced Longformer and SLiding Encoder-Decoder \cite{ivgi2023efficient} to handle long legal documents, while \citet{liu-etal-2022-augmenting} incorporated case-based contrastive embedding enhancement. Nevertheless, these methods focus on picking up the token-level clues instead of thinking based on facts. 

Recently, with the rise of LLMs, carefully designed legal prompts have been employed for judicial decision-making \cite{trautmann2022legal}. ChatLaw \cite{cui2024chatlaw} utilized multi-agent collaboration, while \citet{santosh2024incorporating} incorporated precedents as judgment references to enhance reliability. These methods have been adapted for various regions and languages, including China \cite{sun2024chinese}, Europe \cite{santosh2024incorporating}, Korea \cite{hwang2022multi}, etc \cite{millan2023legal,masala2024improving}. Despite LLMs placing greater emphasis on reasoning, frequent logical errors lead to incorrect final predictions.

\subsection{LLMs Reasoning and Verification}
Although LLMs have demonstrated excellent potential in handling complex tasks, the accumulation of reasoning errors remains a serious challenge. \citet{lightman2023letsverifystepstep} showed that models trained using process rewards outperform those supervised by outcome rewards. To automatically annotate training datasets for process rewards, Math-Shepherd \cite{wang2024mathshepherdverifyreinforcellms} and ReST-MCTS \cite{zhang2024restmctsllmselftrainingprocess} defined it based on their potential to deduce the correct final answer.

To address reasoning errors, researchers have explored LLM self-correction based on output feedback. SELF-REFINE \cite{madaan2024self} introduced an iterative self-improvement strategy, while REFINER \cite{paul-etal-2024-refiner} and $Re^3$ \cite{yang2022re3generatinglongerstories} employed dedicated critic models to provide step-by-step feedback. For code generation, CRITIC \cite{gou2023critic} leveraged external verification tools and fed their critiques back to LLMs. However, these process supervision approaches have mainly been applied to mathematical and coding tasks, and cannot adapt to the LJP task.

\section{Task Definition}
Given a legal case $c$ consisting of a plaintiff's claim $c_{claim}$ and factual descriptions $c_{fact}$, we aim to perform judicial reasoning $\{s_i\}_{i=1}^n$, where each step $s_i$ represents a distinct reasoning component. The goal is to generate judgments $\{j_k\}_{k=1}^m$ covering all judgment items provided by the court and ultimately summarize a final decision $d \in \{0,1\}$, where $d=1$ indicates support for the plaintiff's claim and $d=0$ indicates rejection.

\section{Methodology}
To achieve LJP, LegalReasoner first identifies dispute points to decompose the core issues at hand, then conducts step-by-step reasoning to reach a judgment. During the reasoning process, a process verifier is employed to continuously monitor and validate the logic of each reasoning step. When reasoning flaws are detected, the correction component analyzes the nature of the error and implements targeted solutions to ensure the reliability of the final judgment.

\subsection{Dataset Construction} \label{data_construction}
To enable LLMs to learn dispute identification and conduct reasoning centered around these disputes, we first need to construct a suitable dataset.

\textbf{Corpus Source.} We collected 185,495 Hong Kong judicial cases\footnote{https://www.hklii.hk/databases} from 1900 to 2023, encompassing ten types of judicial decisions: \textit{Court of First Instance, Competition Tribunal, Family Court, Magistrates' Courts}, etc. We excluded cases from the \textit{Court of Appeal} and \textit{Court of Final Appeal} as their official case descriptions are often abbreviated and incomplete.

\textbf{Data Compression and Structuring.} Considering there are extensive lengthy judicial documents, we first performed data compression \cite{pan2024llmlingua}, removing redundant characters while fully preserving semantics. The processed documents were reduced to approximately one-third to one-half of their original length. Next, we employed GPT-4 to extract structured information from judicial documents as depicted in Figure \ref{fig:dataset}. 

\textbf{Refinement and Filtering.} To ensure data quality, we implemented a three-fold refinement and filtering process: 1) We filtered out cases with missing essential components such as plaintiff's claim, dispute points, or legal judgment. 2) To ensure the fact description contained sufficient knowledge to produce judgments consistent with the court's decisions, we augmented it using GPT-4 based on the original document, extracted court reasoning, and legal judgment. 3) To prevent the fact description from revealing the judgment results prematurely, we conducted a thorough verification using both GPT-4 and manual review by judicial experts. After this comprehensive filtering process, the final dataset consisted of 58,130 cases. 

Based on the extracted dispute points, stepwise reasoning, and final judgments from our dataset, we can fine-tune LLMs to learn a judicial reasoning approach that first decomposes cases around key disputes before conducting detailed legal analysis and reaching decisions.

\begin{figure*}[t]
\centering
\begin{minipage}[t]{1\linewidth}
\centering
\includegraphics[width=1.0\textwidth]{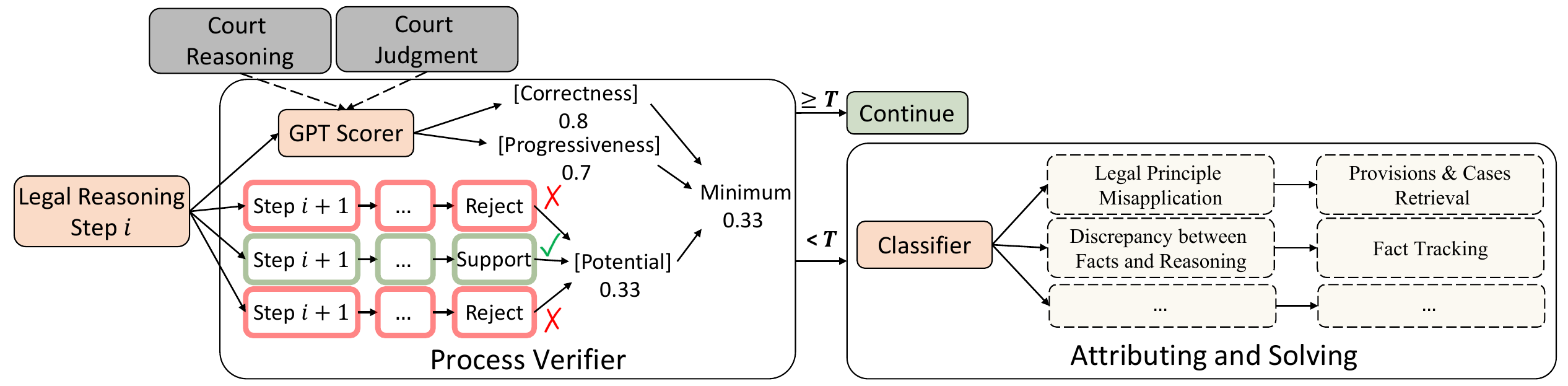}
\end{minipage}
\centering
\caption{Automatic annotation for step-wised verification-correction.}
\label{fig:SWVC}
\end{figure*}
\subsection{Step-wised Verification-Correction} \label{PRM}
To enhance robustness and prevent the accumulation of errors during judicial reasoning, supervision of each step is crucial. Inspired by "LLM as a judge" \cite{zheng2023judging} and "verify step by step" approaches \cite{lightman2023letsverifystepstep,wang2024mathshepherdverifyreinforcellms}, which utilize multiple LLMs or train dedicated process verifiers, we note that these methods are primarily designed for mathematics or general question-answering tasks. To adapt to the unique demands of judicial reasoning, we have developed a specialized verification and correction module specifically tailored for legal reasoning at each process step as shown in Figure \ref{fig:SWVC}.

\subsubsection{Process Verifier} The Process Verifier assigns a score to each reasoning step in the sequence $\{s_i\}_{i=1}^n$. It is built upon an LLM foundation, with fine-tuning focused on optimizing probability distributions across designated special tokens through:
\begin{equation}
    \mathcal{L}_{\text{PV}} = -\sum_{i=1}^{n} y_{s_i} \log r_{s_i} + (1-y_{s_i})\log(1-r_{s_i}), \quad (2)
\end{equation}
where $y_{s_i} \in [0,1]$ represents the quality score of the $i$-th reasoning step $s_i$, $r_{s_i}$ is the sigmoid score of $s_i$ assigned by the verifier, and $n$ is the total number of reasoning steps in the sequence $\{s_i\}_{i=1}^n$.

To automate the construction of label data $y_{s_i}$ for training the process verifier, we evaluate each reasoning step from three perspectives: \textit{Correctness, Progressiveness}, and \textit{Potential}. Specifically, Correctness measures whether the current reasoning step aligns with the facts presented in the case and follows proper legal logic, avoiding contradictions with established evidence. Progressiveness represents the effectiveness of advancing the reasoning toward the final judgment. Both metrics are evaluated through specialized prompts that enable GPT-4 to perform comparisons between case facts, annotated court reasoning, and judgment decisions. Potential indicates the capacity to generate correct judgments by sampling N possible judgment outcomes based on the current reasoning sequence $\{s_i\}_{i=1}^k$. The potential score is formally defined as:

\begin{equation}
    \text{Potential}(s_k) = \frac{\sum_{j=1}^N \mathbb{I}(\{s_i\}_{i=1}^k,\hat{s}_{k+1}^j,...,\hat{s}_n^j \rightarrow d^*)}{N}
\end{equation}
where $s_k$ is the current reasoning step being evaluated, $\hat{s}_{k+1}^j,...,\hat{s}_n^j$ denotes the $j$-th sampled completion of the remaining reasoning steps, $\mathbb{I}(\cdot)$ is the indicator function, $d^*$ is the true judgment decision from the court, and $N$ is the total number of samples. Each aspect corresponds to a special token: [Correctness] [Progressiveness], and [Potential].

Finally, we take the minimum score among these three metrics as the final score for the current reasoning step, deliberately avoiding averaging manners that might assign inappropriately high scores to incorrect or ineffective reasoning steps. If the final score falls below a predetermined threshold $T$, the reasoning step will be flagged as erroneous.

\subsubsection{Problem Attributing and Solving}
\label{Problem_Attributing_Solving}
Once the Process Verifier identifies an error in the reasoning process, it is essential to determine the type of error and apply appropriate remediation strategies. 

Under the guidance of legal experts, we categorized common legal reasoning errors into eight types and subsequently designed corresponding correction strategies based on the characteristics of each error type. Here we present three examples, with complete details provided in Appendix \ref{appendix_B}:
\begin{itemize}
\item \textbf{Legal Principle Misapplication}: This error encompasses the inadequate understanding and incorrect application of legal constituent elements, including misinterpretation of essential legal requirements, failure to establish necessary causal relationships, and improper analysis of legal preconditions. The correction strategy involves: (1) retrieval and analysis of relevant statutory provisions and authoritative case precedents to establish a solid legal foundation, (2) systematic verification of whether each constituent element has been properly established through referencing retrieved legal sources, and (3) rigorous examination of the logical connections between legal requirements and factual evidence, ensuring alignment with established legal interpretations.
\item \textbf{Discrepancy between Facts and Reasoning}: This error occurs when reasoning conclusions lack sufficient factual support or contradict case facts, commonly resulting from overlooking crucial evidence, making unfounded assumptions, or selectively interpreting facts. The correction strategy employs fact tracking to ensure reasoning validity: (1) establishing explicit links between current reasoning steps and their corresponding factual basis in case materials, (2) validating the sufficiency and relevance of referenced facts for supporting the current inference, and (3) revising the reasoning step to maintain strict alignment with tracked factual evidence.
\item \textbf{Errors in Compensation Scope}: This type of error involves incorrect determination of compensation amounts, including improper calculation methods, overlooked compensation items, or exceeding legal compensation limits. The correction strategy implements a structured validation framework: (1) decomposing claimed compensation into atomic components and validating each against specific legal provisions, (2) applying Python scripts to ensure each component's calculation adheres to statutory formulas and prescribed limits, and (3) cross-referencing similar cases to benchmark compensation ratios and identifying potential outliers that require additional justification or adjustment.
\end{itemize}

\section{Experiments}
In this section, we first introduce the experimental setup, and then conduct extensive empirical studies and show the effectiveness of our LegalReasoner.
\subsection{Experiment Setup}
\subsubsection{Datasets}
In addition to our constructed LegalHK dataset, we utilized the latest CMDL dataset \cite{huang2024cmdl}, both of which underwent refinement and filtering processes as described in Section \ref{data_construction}. The LegalHK dataset is derived from Hong Kong court cases spanning from 1900 to 2023, encompassing both civil and criminal cases. It was filtered from 185,495 to 58,130 cases, covering 10 case types and 827 law articles. The CMDL dataset is a Chinese criminal case dataset, sourced from China Judgments Online\footnote{https://wenshu.court.gov.cn/} over the past 20 years. It was filtered from 699,263 to 393,945 cases, containing nearly 1.2 million defendants and involving 321 charges and 275 law articles. We split both datasets into training (50\%), validation (10\%), and test (40\%) sets to ensure robust model evaluation.

\subsubsection{Metrics}
To evaluate our model's performance on LPJ, we employ metrics at two granularity levels. At the case level, we use Case-Level Accuracy (CL-Acc) and Case-Level F1 (CL-F1) to assess the consistency between predicted support/reject decisions and actual court decisions. At the element level, we introduce Element Coverage (E-Cov) to measure the proportion of court judgment elements successfully captured in the model's output, and Element Precision (E-Pre) to quantify the accuracy of the judgment elements when compared against the official court ruling. To compute E-Cov and E-Pre, we use GPT-4 to assess the inclusion of court judgment elements and accuracy of predicted elements. Manual verification confirms that GPT-4's assessments align closely with human evaluation. Predictions of fines are deemed correct if they match the true amount's order of magnitude.

\begin{table*}[t]
\small
\begin{center}
\begin{tabular}{lcccc|cccc} 
\hline
\multirow{2}{*}{\textbf{Model}} & \multicolumn{4}{c|}{\textbf{LegalHK}} & \multicolumn{4}{c}{\textbf{CMDL}} \\
& \textbf{CL-Acc} & \textbf{CL-F1} & \textbf{E-Cov} & \textbf{E-Pre} & \textbf{CL-Acc} & \textbf{CL-F1} & \textbf{E-Cov} & \textbf{E-Pre} \\
\hline
\multicolumn{9}{c}{\textit{BERT-based methods}} \\
NeuralJudge & 63.52 & 61.27 & 13.62 & 14.41 & 57.48 & 55.39 & 12.85 & 13.56 \\
LegalBERT & 64.23 & 61.92 & 14.08 & 14.87 & 58.31 & 56.24 & 13.28 & 13.97 \\
JurBERT & 63.91 & 61.63 & 13.84 & 14.71 & 57.96 & 55.92 & 13.14 & \underline{14.32} \\
Contrastive-LJP & \underline{65.34} & \underline{62.53} & 14.37 & \underline{15.16} & \underline{59.23} & \underline{56.93} & 13.08 & 14.27 \\
ML-LJP & 64.92 & 62.21 & \underline{14.40} & 15.02 & 58.83 & 56.52 & \underline{13.42} & 14.12 \\
\hline
\multicolumn{9}{c}{\textit{LLM-based methods}} \\
LLAMA-3.1-8B & 61.59 & 21.05 & 32.23 & 28.42 & 58.32 & 20.14 & 20.21 & 21.33 \\
\quad + Fine-tuning & 64.47 & 58.52 & 45.48 & 46.52 & 60.27 & 55.64 & 43.61 & 44.15 \\ 
LLAMA-3.1-70B & 70.23 & 67.75 & 52.34 & 53.57 & 65.96 & 63.42 & 50.25 & 51.39 \\
\quad + Fine-tuning & 72.37 & 68.74 & 54.58 & 55.70 & 68.21 & 64.54 & 52.13 & 53.26 \\ 
Qwen-2.5-7B & 66.01 & 53.29 & 45.86 & 46.91 & 61.81 & 50.22 & 43.73 & 44.87 \\
\quad + Fine-tuning & 71.42 & 64.70 & 51.34 & 52.45 & 67.32 & 60.54 & 49.12 & 50.27 \\ 
Qwen-2.5-14B & 69.84 & 67.30 & 49.96 & 51.07 & 65.42 & 63.24 & 47.82 & 48.93 \\
\quad + Fine-tuning & 71.74 & 67.21 & 54.15 & 55.25 & 68.58 & 63.14 & 51.93 & 53.04 \\ 
Qwen-2.5-72B & 70.75 & 69.55 & 53.84 & 54.95 & 66.52 & 65.42 & 51.63 & 52.74 \\
\quad + Fine-tuning & 72.99 & 69.68 & 55.15 & 55.26 & 68.82 & 65.53 & 52.84 & 53.95 \\ 
ChatGPT & 71.43 & 67.86 & 53.93 & 55.04 & 66.35 & 63.72 & 51.84 & 52.95 \\ 
GPT-4 & 73.53 & 70.92 & 61.15 & 62.25 & 68.42 & 66.83 & 57.93 & 57.04 \\ 
Claude-3.5-Sonnet & 73.80 & 70.93 & 62.83 & 62.96 & 68.74 & 67.83 & 58.65 & 57.71 \\
GPT-o1 & \underline{77.15} & \underline{72.13} & \underline{62.92} & \underline{63.14} & \underline{72.03} & \underline{68.35} & \underline{59.54} & \underline{59.12} \\
Dependency-LJP & 66.61 & 60.27 & 47.40 & 48.49 & 62.64 & 57.22 & 45.01 & 46.43 \\
Precedents-LJP & 67.83 & 61.94 & 48.82 & 49.13 & 63.32 & 58.43 & 46.62 & 47.94 \\
\hline
\multicolumn{9}{c}{\textit{Our method}} \\
LegalReasoner & \textbf{80.27} & \textbf{76.33} & \textbf{64.59} & \textbf{65.41} & \textbf{76.95} & \textbf{72.04} & \textbf{62.20} & \textbf{62.36} \\ 
\hline
\end{tabular}
\caption{Overall performance on LegalHK and CMDL datasets. \textbf{Bold} numbers indicate the best score across all models, while underlined numbers represent the best score within each category.}
\label{table:overall_performance}
\end{center}
\end{table*}

\subsubsection{Baselines}
We conduct experiments on both BERT-based and LLM-based methods. For BERT-based methods:
\begin{itemize}
\item NeuralJudge \cite{chalkidis2019neural} enhances pre-trained BERT with LJP-specific fine-tuning.
\item LegalBERT \cite{chalkidis2020legal} pre-trains BERT on legal documents from scratch.
\item JurBERT \cite{masala2024improving} extends LegalBERT with a Sliding Encoder for improved long-context understanding.
\item Contrastive-LJP \cite{liu-etal-2022-augmenting} employs case triples modeling and relational attention to enhance BERT embeddings.
\item ML-LJP \cite{liu2023ml} integrates contrastive learning and Graph Attention Networks to model law article interactions.
\end{itemize}

For LLM-based methods, we compare against both general and legally specialized LLMs:
\begin{itemize}
\item LLAMA-3.1 and Qwen-2.5 represent state-of-the-art open-source language models.
\item ChatGPT, GPT-4, GPT-o1, and Claude-3.5-Sonnet demonstrate strong performance as proprietary models.
\item Dependency-LJP \cite{huang2021dependency} adopts T5 to model logical dependencies in legal judgments.
\item Precedents-LJP \cite{santosh2024incorporating} leverages relevant precedent cases to enhance prediction accuracy.
\end{itemize}

\subsubsection{Implementation Details}
For all BERT-based baselines and open-source LLMs, we fine-tuned them on the LegalHK and CMDL datasets, using plaintiffs' claims and fact descriptions as input and Legal Judgments as labels. For legally specialized LLMs, we utilized LLAMA-3.1-70B as the foundation model along with carefully designed prompts following \citet{trautmann2022legal}.

LegalReasoner consists of two components: a Reasoner and a Verifier, both based on LLAMA-3.1-70B. The Reasoner is fine-tuned on Disputes and Step-wise Reasoning data from LegalHK. The Verifier is fine-tuned using error detection and attribution data annotated as described in Section \ref{PRM}, with detailed training examples provided in Appendix \ref{appendix_A}. We conducted the fine-tuning process using the Llama-factory framework \cite{zheng2024llamafactory}, employing the AdamW optimizer with a learning rate of 1e-5 and weight decay of 0.01, along with a Cosine learning rate schedule. The training epoch is set to 3. The threshold $T$ in process verifier is 0.5. In the correction module, we employ BGE\footnote{https://huggingface.co/BAAI/bge-m3} as the case retriever, utilize LLAMA-3.1-8B with dictionary tree pruning for hierarchical statutory provision retrieval, and use LLAMA-3.1-70B for both reflection and Python script generation.

\subsection{Experimental Results}
Table \ref{table:overall_performance} presents the performance comparison of various methods on LegalHK and CMDL. Among BERT-based methods, Contrastive-LJP achieves the best performance on most metrics for both datasets, with Case-Level Accuracy of 65.34 and 59.23 on LegalHK and CMDL respectively. The performance gap between different BERT-based models is relatively small, suggesting that these architectures may have reached their capacity limits for legal judgment prediction. The notably low E-Cov and E-Pre scores (around 13-15 for all BERT-based methods) indicate their limited ability to capture and analyze detailed judgment elements, revealing significant challenges in fine-grained legal analysis. 

\begin{table*}[t]
\small 
\begin{center}
\begin{tabular}{lcccc|cccc}
\hline
\multirow{2}{*}{\textbf{Model Variant}} & \multicolumn{4}{c|}{\textbf{LegalHK}} & \multicolumn{4}{c}{\textbf{CMDL}} \\
& \textbf{CL-Acc} & \textbf{CL-F1} & \textbf{E-Cov} & \textbf{E-Pre} & \textbf{CL-Acc} & \textbf{CL-F1} & \textbf{E-Cov} & \textbf{E-Pre} \\
\hline
LLAMA-3.1-70B & 72.37 & 68.74 & 54.58 & 55.70 & 68.21 & 64.54 & 52.13 & 53.26 \\
LegalReasoner w/o Disputes & 75.95 & 73.82 & 61.92 & 63.84 & 73.08 & 68.93 & 59.45 & 59.32 \\
LegalReasoner w/o SWR & 69.49 & 65.27 & 54.31 & 55.23 & 66.84 & 62.58 & 52.15 & 53.02 \\
LegalReasoner w/o VC & 74.49 & 71.36 & 56.47 & 58.39 & 71.00 & 67.87 & 55.08 & 56.95 \\
LegalReasoner w/ LLM-Verifier & 78.12 & 74.69 & 61.84 & 62.76 & 74.31 & 70.18 & 59.45 & 60.32 \\
LegalReasoner w/ Judge-Verifier & 78.40 & 74.87 & 62.42 & 63.34 & 74.59 & 70.76 & 60.03 & 60.90 \\
\hline
LegalReasoner (Full) & \textbf{80.27} & \textbf{76.33} & \textbf{64.59} & \textbf{65.41} & \textbf{76.95} & \textbf{72.04} & \textbf{62.20} & \textbf{62.36} \\
\hline
\end{tabular}
\caption{Ablation study results showing the impact of different components in LegalReasoner, where SWR represents Step-Wise Reasoning, VC represents Verification-Correction mechanism, and w/o denotes without.}
\label{table:ablation}
\end{center}
\end{table*}

For LLM-based methods, we observe a clear correlation between model size and performance, with larger models consistently outperforming their smaller counterparts. Fine-tuning on the LJP task significantly improves performance across all open-source LLMs, like Qwen-2.5-72B achieving incremental gains after fine-tuning from 70.75 CL-Acc to 72.99 CL-Acc on LegalHK. Among proprietary models, GPT-o1 demonstrates superior performance, achieving 77.15 CL-Acc on LegalHK and 72.03 on CMDL, demonstrating its strong reasoning capabilities in analyzing complex legal scenarios.

Our proposed LegalReasoner outperforms all baseline methods across both datasets and all metrics. Compared to the strongest baseline (GPT-o1), it achieves gains of 3.12 in CL-Acc and 5.20 in CL-F1 on LegalHK, with even larger improvements of 4.92 in CL-Acc and 3.69 in CL-F1 on CMDL. The superior performance can be attributed to two key factors: the Reasoner component's dispute identification and step-wise reasoning approach that enables structured legal analysis, and the Verifier component that helps identify and correct potential errors. This effectiveness is particularly reflected in the element-level metrics, where we see improvements of 2.67 in E-Cov and 3.27 in E-Pre on LegalHK. The stronger performance on CMDL further demonstrates our model's capability in handling complex criminal cases. Even when using open-source models, we can achieve performance that surpasses GPT-o1.

\subsection{Ablation Studies}
To systematically evaluate the effectiveness of different components in LegalReasoner, we conduct comprehensive ablation studies by creating several model variants and analyzing their performance on both datasets:
\begin{itemize}
\item LLAMA-3.1-70B: The baseline model fine-tuned on the LJP task.
\item LegalReasoner w/o Disputes: Removes the dispute identification component.
\item LegalReasoner w/o SWR: Removes step-wise reasoning, generating judgments directly.
\item LegalReasoner w/o VC: Removes verification-correction mechanism.
\item LegalReasoner w/ LLM-Verifier: Replaces our Process Verifier with direct LLM prompting for verification.
\item LegalReasoner w/ Judge-Verifier: Substitutes our Process Verifier with the "LLM as judge" approach \cite{zheng2023judging}.
\end{itemize}

As shown in Table \ref{table:ablation}, removing dispute identification significantly impairs performance, with CL-Acc falling by 4.32 on LegalHK and 3.87 on CMDL. This demonstrates that decomposing complex cases into key disputes helps the model focus on critical issues and conduct more targeted analysis. The absence of step-wise reasoning also notably impacts performance as CL-Acc drops by 10.78 on LegalHK, indicating its importance in maintaining logical consistency throughout the reasoning process.

Further experiments on verification mechanisms reveal their crucial role in model performance. The removal of our verification-correction mechanism leads to substantial degradation, with CL-Acc declining by 5.78 on LegalHK and 5.95 on CMDL. Alternative approaches also show limitations: the LLM-Verifier variant achieves a 2.15 percentage point lower CL-Acc due to its lack of specialized legal knowledge, while the Judge-Verifier approach exhibits a 1.87 percentage point decrease in CL-Acc. In element-level metrics, E-Cov decreases by 2.17 on LegalHK and 2.47 on CMDL, validating the effectiveness of our process verifier. A more detailed analysis of each correction strategy can be found in Appendix \ref{appendix_C}.

\subsection{Comparison of Verification-Correction and Best of N}
To further validate Verification-Correction approaches, we examine multiple sampling strategies where each case is processed multiple times (N=10), with a verifier selecting the optimal prediction from the generated candidates, called Best of N. We compare several variants: Self-consistency \cite{wang2022self} uses majority voting across reasoning paths, Outcome Verifier only evaluates final judgments, General Process Verifier \cite{lightman2023letsverifystepstep} employs a general reasoning corpus, and Legal Process Verifier utilizes training from LegalHK.

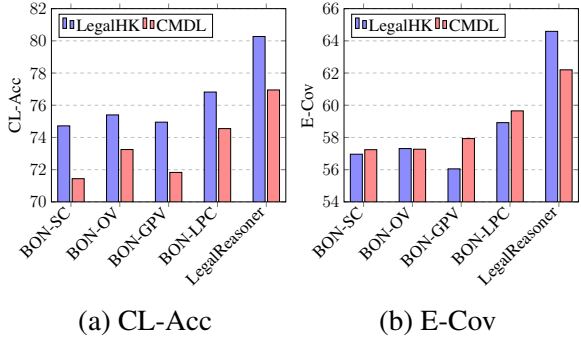
\begin{figure}[t]
    \begin{minipage}[t]{0.5\linewidth}
        \centering
        \begin{tikzpicture}[scale=0.45]
        \begin{axis}[
            ybar,
            ylabel={CL-Acc},
            symbolic x coords={BON-SC,BON-OV,BON-GPV,BON-LPC,LegalReasoner},
            xtick=data,
            xticklabel style={rotate=45, anchor=east},
            legend style={at={(0.36,0.85)}, anchor=south, legend columns=-1},
            ymajorgrids=true,
            grid style=dashed,
            ymin=70,
            ymax=82,
            font=\Large,
        ]
        \addplot[fill=blue!45] coordinates {
            (BON-SC,74.72)
            (BON-OV,75.40)
            (BON-GPV,74.95)
            (BON-LPC,76.82)
            (LegalReasoner,80.27)
        };
        \addplot[fill=red!45] coordinates {
            (BON-SC,71.44)
            (BON-OV,73.25)
            (BON-GPV,71.83)
            (BON-LPC,74.55)
            (LegalReasoner,76.95)
        };
        \legend{LegalHK, CMDL}
        \end{axis}
        \end{tikzpicture}
        \centerline{(a) CL-Acc}
    \end{minipage}%
    \begin{minipage}[t]{0.5\linewidth}
        \centering
        \begin{tikzpicture}[scale=0.45]
        \begin{axis}[
            ybar,
            ylabel={E-Cov},
            symbolic x coords={BON-SC,BON-OV,BON-GPV,BON-LPC,LegalReasoner},
            xtick=data,
            xticklabel style={rotate=45, anchor=east},
            legend style={at={(0.36,0.85)}, anchor=south, legend columns=-1},
            ymajorgrids=true,
            grid style=dashed,
            ymin=54,
            ymax=66,
            font=\Large,
        ]
        \addplot[fill=blue!45] coordinates {
            (BON-SC,56.96)
            (BON-OV,57.31)
            (BON-GPV,56.05)
            (BON-LPC,58.92)
            (LegalReasoner,64.59)
        };
        \addplot[fill=red!45] coordinates {
            (BON-SC,57.24)
            (BON-OV,57.27)
            (BON-GPV,57.93)
            (BON-LPC,59.65)
            (LegalReasoner,62.20)
        };
        \legend{LegalHK, CMDL}
        \end{axis}
        \end{tikzpicture}
        \centerline{(b) E-Cov}
    \end{minipage}
    \caption{Comparison of Best of N (BON) and our approach, where SC is Self-consistency, OV is Outcome Verifier, GPV is General Process Verifier, and LPC is Legal Process Verifier.}
    \label{fig:verification-comparison}
\end{figure}

Results in Figure \ref{fig:verification-comparison} show LegalReasoner outperforms all Best of N variants, achieving CL-Acc scores of 80.27 on LegalHK and 76.95 on CMDL. The Legal Process Verifier performs best among Best of N approaches with 76.82 on LegalHK and 74.55 on CMDL, while general approaches achieve around 74-75 on LegalHK and 71-72 on CMDL. As shown in Table \ref{table:ablation}, these scores match LegalReasoner's base performance of 74.49 without verification-correction, suggesting that increased sampling without proper error correction offers limited improvement in legal reasoning.


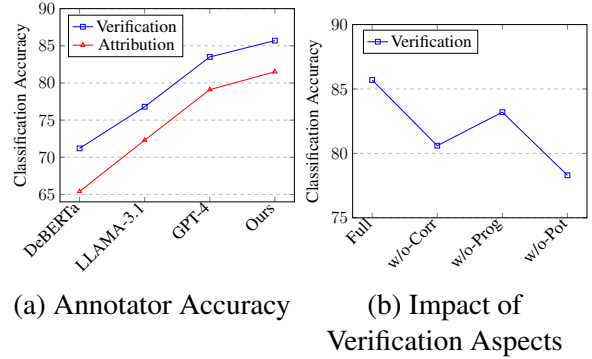
\begin{figure}[t]
    \begin{minipage}[t]{0.5\linewidth}
        \centering
        \begin{tikzpicture}[scale=0.45]
        \begin{axis}[
            ylabel={Classification Accuracy},
            symbolic x coords={DeBERTa,LLAMA-3.1,GPT-4,Ours},
            xtick=data,
            xticklabel style={rotate=45, anchor=east},
            legend style={at={(0.28,0.75)}, anchor=south, legend columns=1},
            ymajorgrids=true,
            grid style=dashed,
            ymin=64,
            ymax=90,
            font=\Large,
        ]
        \addplot[mark=square,blue,thick] coordinates {
            (DeBERTa,71.2)
            (LLAMA-3.1,76.8)
            (GPT-4,83.5)
            (Ours,85.7)
        };
        \addlegendentry{Verification}
        \addplot[mark=triangle,red,thick] coordinates {
            (DeBERTa,65.4)
            (LLAMA-3.1,72.3)
            (GPT-4,79.1)
            (Ours,81.5)
        };
        \addlegendentry{Attribution}
        \end{axis}
        \end{tikzpicture}
        \centerline{(a) Annotator Accuracy}
    \end{minipage}%
    \begin{minipage}[t]{0.5\linewidth}
        \centering
        \begin{tikzpicture}[scale=0.45]
        \begin{axis}[
            ylabel={Classification Accuracy},
            symbolic x coords={Full,w/o-Corr,w/o-Prog,w/o-Pot},
            xtick=data,
            xticklabel style={rotate=45, anchor=east},
            legend style={at={(0.29,0.85)}, anchor=south, legend columns=-1},
            ymajorgrids=true,
            grid style=dashed,
            ymin=75,
            ymax=90,
            font=\Large,
        ]
        \addplot[mark=square,blue,thick] coordinates {
            (Full,85.7)
            (w/o-Corr,80.6)
            (w/o-Prog,83.2)
            (w/o-Pot,78.3)
        };
        \addlegendentry{Verification}
        \end{axis}
        \end{tikzpicture}
        \centerline{(b) Impact of}
        \centerline{Verification Aspects}
    \end{minipage}
    \caption{Assessment of automatic annotations for process verification and attribution, where Corr stands for Correctness, Prog for Progressiveness, and Pot for Potential.}
    \label{fig:annotation-quality}
\end{figure}
\vspace{-0.2cm}
\subsection{Quality of Automatic Annotations}
To explore the quality of our automatic dataset, we manually annotated 160 reasoning steps randomly sampled from the training set of LegalHK. We employed different models to infer labels for each step and compared their performance.

\textbf{Quality of Automatic Annotations.} As shown in Figure \ref{fig:annotation-quality}(a), we evaluated both verification accuracy and attribution accuracy across different models. DeBERTa achieves verification accuracy of 71.2 and attribution accuracy of 65.4. LLAMA-3.1-70B with carefully designed prompts shows improved performance, reaching 76.8 and 72.3 respectively. GPT-4 further enhances the results to 83.5 and 79.1. Our fine-tuned Process Verifier demonstrates the best performance with 85.7 verification accuracy and 81.5 attribution accuracy, indicating that our automatic annotation process achieves satisfactory quality.


\textbf{Impact of Verification Perspectives.} As shown in Figure \ref{fig:annotation-quality}(b), we evaluate each reasoning step from three perspectives: Correctness, Progressiveness, and Potential. The full model achieves 85.7 verification accuracy, while removing each perspective leads to different levels of performance degradation. Correctness proves crucial, with accuracy dropping to 80.6, while accuracy decreases to 78.3 without Potential and to 83.2 without Progressiveness. These results validate the effectiveness of our multi-perspective verification in generating quality training data for LegalReasoner.

\section{Conclusion}
In this paper, we proposed LegalReasoner to enhance the reliability of legal judgment prediction through step-wise verification and correction on the reasoning process. It first identifies dispute points to decompose complex cases and employs a process verifier to validate each reasoning step's logic. For errors detected, expert-designed strategies are applied for correction. We also released the LegalHK dataset with 58,130 annotated Hong Kong court cases for the training of LegalReasoner. Experimental results demonstrated significant improvements over existing methods across multiple datasets.

\section*{Limitations}
Our work has several important limitations that should be acknowledged:

First, the computational cost of our verification-correction process is substantial. While the Process Verifier improves prediction reliability, it requires significant computing resources to validate each reasoning step and implement corrections. This is particularly evident when handling complex cases with multiple dispute points and lengthy reasoning chains. Our approach essentially extends the chain-of-thought (CoT) paradigm by adding verification and correction mechanisms at each step, resulting in longer and more complex reasoning pathways that demand additional computational overhead. However, compared to Best of N sampling approaches which require multiple complete reasoning paths to be generated, our single-path verification-correction method actually achieves better performance with lower computational cost. Although this cost is still lower than human legal analysis, it may limit the method's applicability in resource-constrained environments.

Second, the automatic annotation process for training the Process Verifier contains inherent noise. While our experiments demonstrate the effectiveness of our approach, the quality of verification labels generated through automated means may not fully match human expert annotations. This is particularly relevant for the scoring of reasoning steps from correctness, progressiveness, and potential perspectives. A comprehensive comparison between human and automated annotations could provide valuable insights for future improvements.

Third, while our model shows strong performance in the Chinese legal system, its generalizability to other jurisdictions requires further investigation. Different legal systems may have varying approaches to case analysis and judgment formation, which could affect the effectiveness of our dispute identification and reasoning verification mechanisms. Future work could explore adapting our approach to diverse legal frameworks and validating its performance across different jurisdictions.

\section*{Ethics Statement}
Our work utilizes publicly available court judgments from Hong Kong and Chinese court databases, which have undergone official anonymization processes by their respective judicial authorities. Personal information in these cases has been handled according to established judicial privacy protocols: private individuals' information is anonymized in the source documents, while information about public figures and companies remains unchanged as per standard legal practice. We do not perform additional anonymization beyond these official procedures.

It is crucial to acknowledge that training models on historical legal decisions may perpetuate existing biases in the judicial system. While our model shows improved accuracy in legal reasoning, it inherits biases both from the historical data and the pre-trained language models used as its foundation. This could potentially lead to disparities in performance across different demographic groups or case types.

We emphasize that LegalReasoner is developed as a research tool to advance our understanding of legal reasoning in AI systems, not as a replacement for human judicial decision-making. The legal judgment prediction (LJP) task raises significant ethical and legal concerns, particularly regarding the role of automated systems in judicial processes. We do not advocate for the direct deployment of LJP systems in courts. Instead, our research aims to explore how AI systems can better understand and process legal reasoning patterns while maintaining transparency and accountability.

The step-wise verification and correction mechanisms in LegalReasoner are designed to make the reasoning process more transparent and analyzable, allowing for better scrutiny of potential biases and errors. However, these mechanisms should be viewed as technical contributions toward more reliable and explainable legal AI systems, rather than as solutions for automated judicial decision-making.

Our research group remains committed to developing AI systems that can assist legal professionals while maintaining high standards of ethical responsibility, transparency, and fairness. We encourage future research to further investigate the ethical implications of legal AI systems and develop more robust methods for bias detection and mitigation.

\section*{Acknowledgments}
This work is funded in part by the HKUST Start-up Fund (R9911), Theme-based Research Scheme grant (No. T45-205/21-N), the InnoHK funding for Hong Kong Generative AI Research and Development Center, Hong Kong SAR, the National Natural Science Foundation of China (Grant No. 62102277), the National Key R\&D Program of China under Grant No. 2022YFC3303600, and the Zhejiang Provincial Natural Science Foundation of China under Grant No. LY23F020010.

\bibliography{custom}

\appendix

\section{Dataset Examples and Annotation Details}
\label{appendix_A}
\subsection{Dataset Construction Process}
The construction of LegalHK dataset involved a systematic multi-stage process beginning with comprehensive data collection from the Hong Kong Legal Information Institute (HKLII) database. The initial corpus encompassed 185,495 cases spanning from 1900 to 2023, drawn from the complete hierarchy of Hong Kong's judicial system. The dataset includes cases from fourteen distinct judicial bodies:

The Court of Final Appeal (HKCFA) represents the highest court, followed by cases from the United Kingdom Privy Council Judgments for Hong Kong (UKPC) from the colonial period. The Court of Appeal (HKCA) and Court of First Instance (HKCFI) form the High Court level. The specialized tribunals include the Competition Tribunal (HKCT), Labour Tribunal (HKLaT), and Lands Tribunal (HKLdT). At the lower courts level, we have the District Court (HKDC), Family Court (HKFC), and Magistrates' Courts (HKMagC). Additionally, the dataset encompasses specialized judicial bodies including the Coroner's Court (HKCrC), Small Claims Tribunal (HKSCT), and Obscene Articles Tribunal (HKOAT).

We initially excluded cases from the Court of Final Appeal and Court of Appeal from our analysis set due to their tendency to present abbreviated case descriptions that lack detailed reasoning. However, these cases were retained in our reference database as they often contain important precedential value and legal principles that inform lower court decisions.

The first stage of processing focused on document compression while maintaining semantic integrity. We employed LLM-based compression techniques to reduce document length to approximately one-half to one-third of the original size. This compression preserved all essential legal content while eliminating redundant language and repetitive procedural descriptions that did not contribute to the core legal analysis.

Quality control formed a crucial component of our dataset construction methodology. We first applied a rigorous completeness check that filtered out cases missing any essential components such as plaintiff's claims, clear dispute points, or final legal judgments. This initial filtering significantly reduced our corpus to 58,130 cases, each containing complete information necessary for meaningful legal analysis.

\subsection{Fact Description Enhancement and Verification}
\begin{table*}[t]
\centering
\begin{tabular}{p{0.15\textwidth}p{0.8\textwidth}}
\toprule
\textbf{Stage} & \textbf{Prompt Template} \\
\midrule
Initial Extraction & 
Please extract the following key information from the given court judgment and return it in JSON format:
\begin{verbatim}
{
    "plaintiff": "Name of the applicant/plaintiff",
    "defendant": "Name of the defendant",
    "plaintiff_claim": "claims made by the plaintiff",
    "lawsuit_type": "e.g., judicial review application",
    "facts": ["List of detailed factual background information"],
    "related_laws": ["List of relevant laws or legal provisions"],
    "relevant_cases": ["List of relevant case law or precedents cited"],
    "issues": ["List of main issues in dispute (if any)"],
    "court_reasoning": ["List of reasoning process or legal principles"],
    "judgment_decision": ["List of court's final decision and orders"]
}
\end{verbatim}

Requirements:
1. Clearly distinguish between factual statements and court opinions
2. Organize facts and procedural content chronologically
3. Return empty string "" for fields where information is not explicit
4. Exhibit detailed information for facts and court reasoning
5. Maintain JSON format integrity and standards \\
\midrule
Issue Generation & 
Please generate disputes about plaintiff claim based on facts. These disputes are vital to determine whether Courts support or reject the application/appeal.

plaintiff\_claim: [plaintiff's claims]
facts: [list of facts] \\
\midrule
Fact Enhancement & 
Please review the court's reasoning section and extract any additional factual information to supplement the existing facts. Return the results in JSON format:
\begin{verbatim}
{
    "plaintiff": "...",
    "defendant": "...",
    "plaintiff_claim": "...",
    "lawsuit_type": "...",
    "more_facts": ["List of detailed factual information"],
    "related_laws": ["..."],
    "relevant_cases": ["..."],
    "issues": ["..."],
    "court_reasoning": ["..."],
    "judgment_decision": ["..."],
    "support&reject": "support/reject"
}
\end{verbatim}

Requirements:
1. Review court\_reasoning to identify:
   - Facts mentioned in decision explanation
   - Facts referenced in evidence analysis
   - Facts cited when applying legal principles
   - Facts supporting final judgment
2. Extract only objective factual statements
3. Add new facts to more\_facts if not in original facts
4. Maintain chronological order
5. Do not modify other sections \\
\bottomrule
\end{tabular}
\caption{Prompts for fact description enhancement and verification.}
\label{tab:prompts}
\end{table*}
The fact description sections underwent a specialized enhancement process to ensure they contained sufficient information for reproducing court judgments as shown in Table \ref{tab:prompts}. This enhancement drew upon three primary sources: the original court documents, extracted court reasoning chains, and final legal judgments. The enhanced descriptions were carefully crafted to provide comprehensive context while maintaining neutrality and avoiding premature revelation of case outcomes.

To validate the enhanced fact descriptions, we implemented a two-tier verification system. The first tier utilized GPT-4 for automated screening to identify potential information leakage or bias in the descriptions. The second tier involved a manual review by legal experts who assessed both the completeness and neutrality of the enhanced content. Cases where fact descriptions could potentially telegraph the outcome were either revised or removed from the dataset.

\begin{figure*}[t]
\centering
\begin{minipage}[t]{1\linewidth}
\centering
\includegraphics[width=1.0\textwidth]{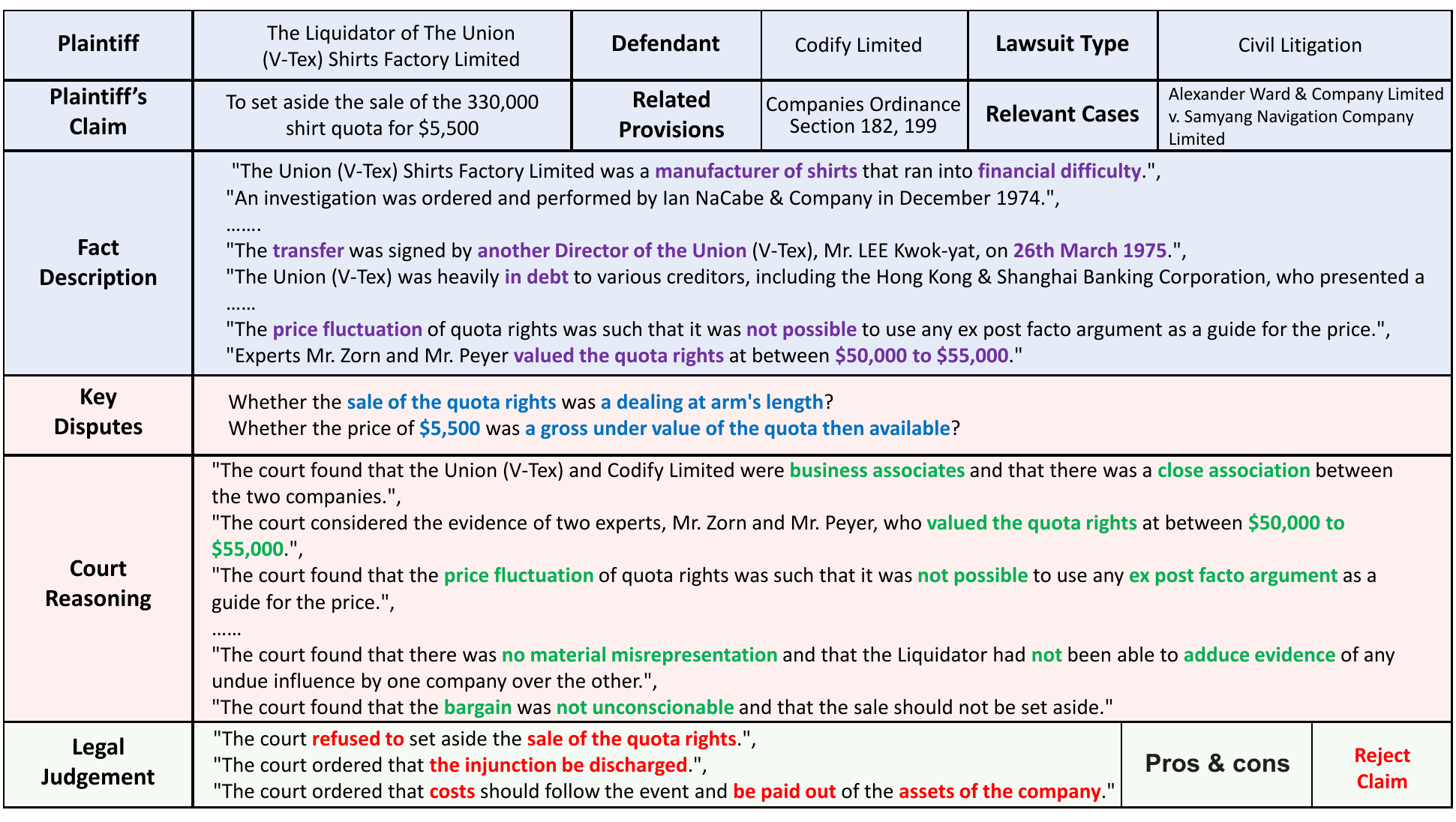}
\end{minipage}
\centering
\caption{Illustration of LegalHK: an annotated legal judgment dataset with disputes and reasoning.}
\label{fig:dataset}
\end{figure*}

\subsection{Annotation Schema and Example Analysis}
Our annotation schema was designed to capture the full complexity of legal decision-making while maintaining consistent structure across diverse case types. Taking the example case illustrated in Figure \ref{fig:dataset}, we can observe how this schema operates in practice. The case involves The Liquidator of The Union (V-Tex) Shirts Factory Limited seeking to set aside a sale of 330,000 shirt quota for \$5,500 to Codify Limited.

The fact description captures several crucial elements: the company's status as a shirt manufacturer facing financial difficulties, the timing of an investigation in December 1974, the transfer's execution by Director LEE Kwok-yat on March 26, 1975, the company's debt situation, and expert valuations placing the quota rights between \$50,000 and \$55,000. These facts are presented chronologically and with clear attribution to establish a comprehensive factual foundation.

The dispute points section distills the case into two fundamental questions: whether the sale represented an arm's length transaction and whether the \$5,500 price constituted a gross undervaluation of the quota rights. This distillation helps focus the subsequent legal analysis on the core issues requiring judicial determination.

The court's reasoning follows a structured progression through several key considerations. It begins by examining the business relationship between the parties, evaluates expert testimony regarding valuation, analyzes price fluctuation patterns, investigates potential misrepresentation, and assesses the overall fairness of the bargaining process. This reasoning chain demonstrates the step-by-step logical progression typical of judicial decision-making.

\begin{figure*}[th]
\centering
\begin{minipage}[t]{1\linewidth}
\centering
\includegraphics[width=1.0\textwidth]{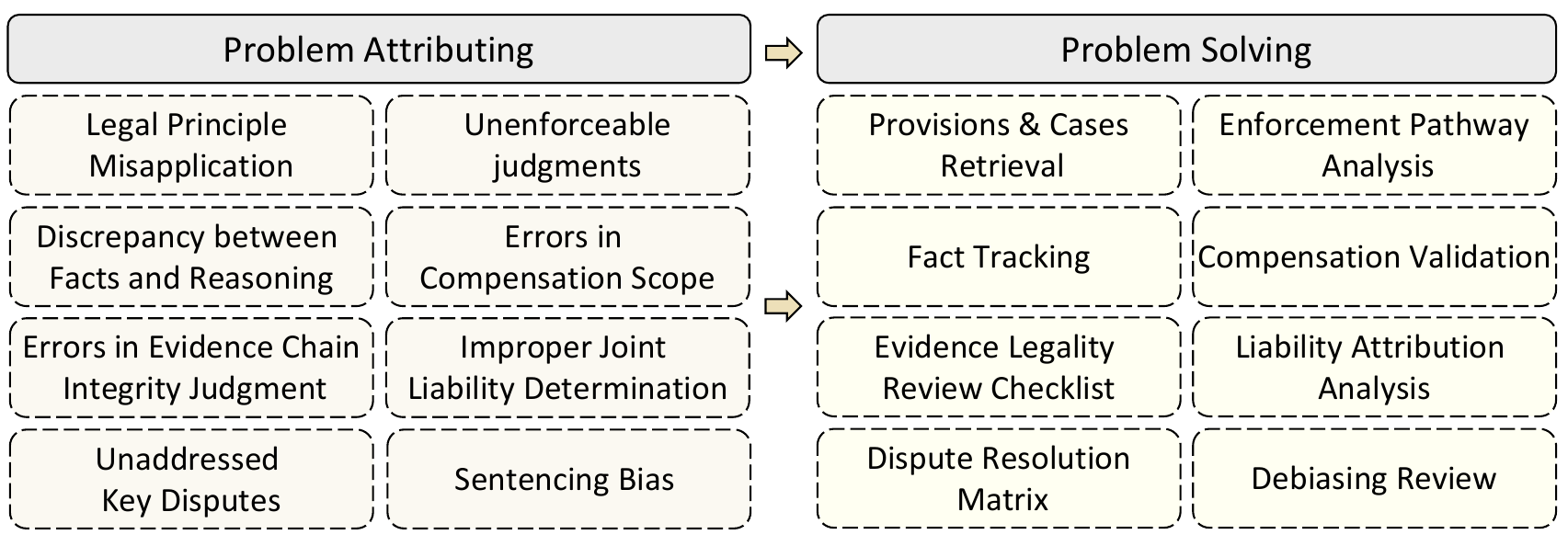}
\end{minipage}
\centering
\caption{Overview of error types and corresponding correction strategies in LegalReasoner. The framework consists of two main components: Problem Attributing (left) which identifies eight distinct types of legal reasoning errors, and Problem Solving (right) which implements specialized correction strategies for each error type.}
\label{fig:error_strategy}
\end{figure*}

\subsection{Quality Assurance and Validation}
Our quality assurance process employed multiple validation mechanisms to ensure annotation consistency and accuracy. Rather than relying solely on individual annotator judgment, we implemented a consensus-based review system where multiple legal experts examined each annotation. Disagreements were resolved through structured discussion to establish consistent annotation standards across the dataset. Notice that all annotators were invited or volunteered to participate and were undergraduate students majoring in law.

The verification process included automated consistency checks using GPT-4, followed by targeted manual sampling to verify accuracy. We maintained detailed error rate monitoring and implemented immediate correction procedures when systematic issues were identified. This multi-layered approach helped ensure the reliability and reproducibility of our annotations while maintaining the nuanced understanding necessary for legal analysis.

Completeness validation formed the final stage of our quality assurance process. Each case underwent systematic checking for component completeness, cross-referencing with original documents, and validation of logical flow. This thorough approach helped ensure that our final dataset maintains high standards of accuracy and completeness while preserving the complex legal reasoning essential for judicial decision-making.

\section{Error Types and Correction Strategies}
\label{appendix_B}
Building upon the three examples introduced in Section \ref{Problem_Attributing_Solving}, we present the remaining five error types and their corresponding correction strategies shown in Figure \ref{fig:error_strategy}. These strategies primarily rely on three implementation approaches: retrieval of external information like statutory provisions, LLM-based self-reflection and adjustment through carefully designed prompts, and evidence tracking to ensure consistency.

\textbf{Unenforceable Judgments} occur when legal conclusions cannot be effectively executed due to practical limitations or jurisdictional issues. The correction process begins with retrieval of enforcement regulations and precedents to understand execution requirements. LLM-based reflection then evaluates practical feasibility and suggests alternative approaches while maintaining legal intent. Finally, consistency validation ensures the modified judgment aligns with established enforcement mechanisms.

\textbf{Errors in Evidence Chain Integrity Judgment} occurs when the analysis of evidence reliability and logical connections is flawed. The correction strategy employs a structured approach that systematically traces each piece of evidence cited in the reasoning process. Through careful prompt-based analysis, it examines both individual evidence reliability and interconnections between different pieces of evidence. The strategy evaluates factors such as credibility, relevance, and logical coherence, ensuring that conclusions drawn from the evidence chain are properly supported and logically sound.

\textbf{Improper Joint Liability Determination} represents errors in assigning responsibility among multiple parties. The strategy first retrieves relevant joint liability provisions and precedents. Through structured prompts, the LLM maps relationships between parties and analyzes their respective roles. Consistency checking ensures liability distribution aligns with both legal principles and factual evidence.

\textbf{Unaddressed Key Disputes} arise when significant legal issues raised by parties are overlooked in the reasoning process. The correction strategy begins by systematically identifying and extracting all disputes mentioned in the case materials, including both explicit claims and implicit contentions. Through structured prompting, it evaluates the current reasoning's coverage of each dispute and ensures appropriate analytical depth is applied to each issue. 

\textbf{Sentencing Bias} manifests when penalties deviate from standards without justification. The strategy retrieves a database of similar cases and sentencing guidelines. Prompt-based analysis examines aggravating and mitigating factors, while consistency validation ensures penalties align with both legal principles and case-specific circumstances.

This implementation framework combines information retrieval, LLM-based reasoning, and consistency validation to systematically identify and address potential flaws in legal analysis. Each strategy leverages these approaches to not only correct immediate errors but also strengthen the overall reliability of legal reasoning.

\begin{figure}[t]
\centering
\begin{tikzpicture}[scale=0.95]
\begin{axis}[
    ylabel={CL-Acc},
    symbolic x coords={Base,LP,FD,CS,UE,EC,JL,KD,SB,LegalReasoner},
    xtick=data,
    xticklabel style={rotate=45, anchor=east},
    ymajorgrids=true,
    grid style=dashed,
    ymin=70,
    ymax=82,
    legend style={at={(0.77,0.07)}, anchor=south, legend columns=1},
]

\addplot[mark=square,blue,thick] coordinates {
    (Base,74.49)
    (LP,77.15)
    (FD,78.83)
    (CS,79.72)
    (UE,79.85)
    (EC,78.91)
    (JL,79.84)
    (KD,78.32)
    (SB,78.76)
    (LegalReasoner,80.27)
};

\addplot[mark=triangle,red,thick] coordinates {
    (Base,71.00)
    (LP,74.15)
    (FD,75.83)
    (CS,76.42)
    (UE,76.65)
    (EC,75.91)
    (JL,76.84)
    (KD,75.32)
    (SB,75.76)
    (LegalReasoner,76.95)
};

\legend{LegalHK, CMDL}
\end{axis}
\end{tikzpicture}
\caption{Impact of different correction strategies on model performance. The base stands for removing all strategies, LP for Legal Principle, FD for Facts Discrepancy, CS for Compensation Scope, UE for Unenforceable, EC for Evidence Chain, JL for Joint Liability, KD for Key Disputes, and SB for Sentencing Bias.}
\label{fig:strategy_effect}
\end{figure}
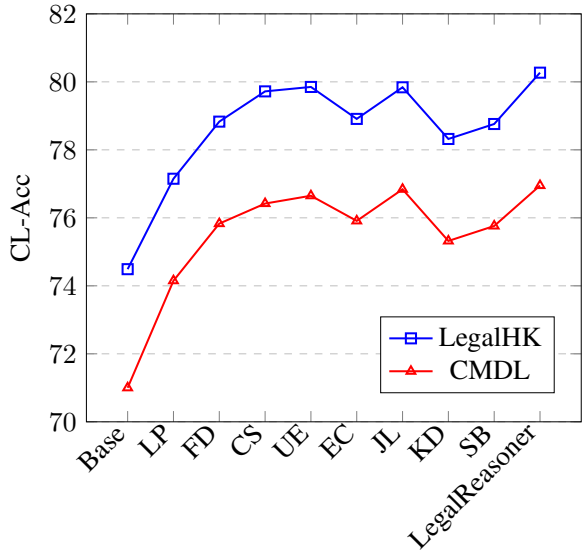
\section{Strategy-wise Effectiveness Analysis}
\label{appendix_C}
To evaluate the contribution of each correction strategy, we conducted ablation experiments by removing individual strategies while keeping others intact. Figure \ref{fig:strategy_effect} illustrates the performance impact of each strategy's removal on both LegalHK and CMDL datasets.

LegalReasoner with all correction strategies achieves the highest performance with a CA-Acc of 80.27 on LegalHK and 76.95 on CMDL. In contrast, the baseline model without any correction strategies performs at 74.49 on LegalHK and 71.00 on CMDL. Ablation results reveal varying performance impacts: removing Legal Principle (LP) strategy reduces performance to 77.15 and 74.15 on LegalHK and CMDL respectively; without Facts Discrepancy (FD), performance drops to 78.83 and 75.83; removing Compensation Scope (CS) leads to 79.72 and 76.42; absence of Unenforceable (UE) results in 79.85 and 76.65; eliminating Evidence Chain (EC) decreases performance to 78.91 and 75.91; without Joint Liability (JL), accuracy falls to 79.84 and 76.84; removing Key Disputes (KD) results in 78.32 and 75.32; and without Sentencing Bias (SB), performance declines to 78.76 and 75.76 respectively.

The consistent performance degradation observed when removing individual strategies across both datasets validates the effectiveness of our comprehensive correction framework. The varying magnitude of performance drops for different strategies suggests that a combination of verification and correction mechanisms is essential for robust legal reasoning.

\section{Retrieval Implementation Details}
For statutory provision retrieval, we constructed a hierarchical directory structure of legal provisions, enabling LLMs to perform progressive pruning through chapter names, part names, article names, section names, and specific provisions. This hierarchical approach significantly improves efficiency by avoiding unnecessary token consumption during the retrieval process and reduces the computational overhead of processing irrelevant legal provisions.

For case retrieval, we established a comprehensive case database on the Dify platform \footnote{https://github.com/langgenius/dify} containing all relevant precedents. We employed the BGE-M3 model\footnote{https://huggingface.co/BAAI/bge-m3} as the primary retriever, followed by BGE-reranker-large\footnote{https://huggingface.co/BAAI/bge-reranker-large} for result reranking. To maintain efficiency while ensuring relevance, we retain only the top-3 ranked cases as supplementary context. It is then summarized and used to enhance the reasoning process and improve the accuracy of legal analysis.

\label{sec:appendix}

\section{Definition of Dispute Points}
\label{app:dispute_definition}

Dispute points constitute the fundamental areas of disagreement between plaintiff and defendant in legal proceedings. To illustrate this concept, we present two paradigmatic examples:

\begin{enumerate}
    \item \textbf{Contract Dispute Case}: A critical dispute point may be formulated as ``\textit{Whether the alleged oral agreement constitutes a binding legal obligation upon the contracting parties}.''
    
    \item \textbf{Bankruptcy Liquidation Case}: A pivotal dispute point might be expressed as ``\textit{Whether the corporate entity possesses sufficient financial capacity to maintain ongoing operations}.''
\end{enumerate}

In both instances, one party maintains an affirmative position while the opposing party presents a contradictory stance. This fundamental disagreement necessitates judicial determination—whether by a presiding judge or an LLM system performing Legal Judgment Prediction (LJP). Once this determination is rendered, the allocation of legal responsibility becomes unambiguous and enforceable.

\section{Runtime Efficiency Analysis}
\label{app:efficiency_analysis}

The computational efficiency analysis of LegalReasoner reveals additional overhead beyond the foundation model's baseline reasoning capabilities. These incremental costs primarily stem from: (1) extended Chain-of-Thought (CoT) reasoning sequences, (2) Verifier classification processes, (3) LLM reflection mechanisms during error correction, and (4) intermittent legal provision and case law retrievals.

Utilizing the vLLM framework, our LLAMA-8B implementation processes approximately 15,000 input tokens and generates 793 output tokens per second on H800 server infrastructure. Consequently, our analysis focuses predominantly on output token generation metrics.

\textbf{Component-wise Token Consumption:}
\begin{itemize}
    \item \textit{Dispute Generation}: 300 tokens (\textasciitilde{}0.4 seconds)
    \item \textit{Judicial Reasoning Process}: 800--1,000 tokens (\textasciitilde{}1.0 second)
    \item \textit{Verifier Classification}: 1 token (\textasciitilde{}0.1 seconds)
    \item \textit{Legal Provision Retrieval}: 2,500 tokens (\textasciitilde{}1.0 second) via 3B LLM for batch directory filtering
    \item \textit{Case Law Retrieval}: Negligible computational cost (\textasciitilde{}0.1 seconds) using 300M BGE model with vector database optimization
    \item \textit{Reflection Process}: 700 tokens (\textasciitilde{}0.9 seconds)
\end{itemize}

This architecture enables comprehensive user response generation within a 3-second timeframe. In comparative analysis with GPT-o1, which employs opaque reasoning processes, our system demonstrates superior efficiency with first-token response times of approximately 7 seconds, while avoiding substantial token expenditure on non-essential descriptive content unrelated to the reflection mechanism.

\section{Case Example}
\label{app:case_study}

The following case study demonstrates the practical application of LegalReasoner's analytical framework:

\begin{tcolorbox}[colback=gray!5!white,colframe=gray!75!black,title=\textbf{Case Study: Chen v. MTR Corporation Limited},breakable]

\textbf{Case Facts:} On 26 October 2009, Ms. Chen (plaintiff, aged 56, employed as editor for ecclesiastical organization) sustained injuries from a slip-and-fall incident within the South Passenger Concourse of Kowloon Tong Mass Transit Railway Station at approximately 09:00 hours during Chung Yeung Festival. Plaintiff alleges surface water presence as causative factor. Meteorological conditions were dry. Station experienced elevated passenger volume consistent with morning rush hour patterns. Defendant MTR Corporation had contracted Winson Cleaning Services as independent contractor for facility maintenance. Systematic inspections occurred bi-hourly. Comprehensive CCTV coverage and appropriate warning signage were maintained. Witness testimonies from Mr. Lau (plaintiff's associate) and Ms. Wong (bystander) presented contradictory accounts regarding floor conditions. Maintenance documentation indicated routine cleaning procedures 30 minutes prior to incident, with no reported spillage events.

\textbf{Plaintiff's Allegations:} Defendant's negligence and breach of common duty of care proximately caused slip-and-fall incident, resulting in substantial lower back and right knee injuries.

\textbf{LegalReasoner Analysis:} Initial assessment identified two principal legal issues: (1) determination of floor surface conditions at incident time, and (2) applicability of statutory defenses under Occupiers Liability Ordinance s. 3(4)(b). Preliminary analysis erroneously applied strict liability principles.

\textbf{Legal Research Triggered:} Occupiers Liability Ordinance, Cap. 314; relevant precedents including \textit{Hsu Li Yun v The Incorporated Owners of Yuen Fat Building} [2000] 1 HKLRD 900, \textit{Cheung Wai Mei v The Excelsior Hotel}, and \textit{Ward v Tesco Stores Ltd.} [1976] 1 WLR 810.

\textbf{Reflection and Correction:} Corrected analysis established that Occupiers Liability Ordinance mandates reasonable care standard rather than strict liability. Evidence demonstrated defendant's compliance through: engagement of qualified contractor, systematic bi-hourly inspections, appropriate warning systems, and comprehensive monitoring protocols. Maintenance records and witness inconsistencies supported defendant's reasonable care demonstration under s. 3(4)(b).

\textbf{Judgment:} Plaintiff's claim dismissed. Costs awarded to defendant with counsel certification, effective 14 days post-judgment.

\textbf{Outcome Classification:} Claim Rejected
\end{tcolorbox}

\end{document}